\documentclass[]{fairmeta}

\title{WorldPrediction: A Benchmark for High-level World Modeling and Long-horizon Procedural Planning}

\author[1,2,*]{Delong Chen}
\author[1,3,*]{Willy Chung}
\author[1,2]{Yejin Bang}
\author[1,2]{Ziwei Ji}
\author[1,2]{Pascale Fung}

\affiliation[1]{Meta FAIR Paris}
\affiliation[2]{The Hong Kong University of Science and Technology}
\affiliation[3]{ISIR Sorbonne Université}

\contribution[*]{Joint first author}

\abstract{
Humans are known to have an internal ``world model'' that enables us to carry out action planning based on world states. AI agents need to have such a world model for action planning as well.  It is not clear how current AI models, especially generative models, are able to learn such world models and carry out procedural planning in diverse environments. We introduce \textsc{WorldPrediction}, a video-based benchmark for evaluating
world modeling and procedural planning capabilities of different AI models. In contrast to prior benchmarks
that focus primarily on low-level world modeling and robotic motion planning, \textsc{WorldPrediction} is the first benchmark that emphasizes actions with temporal and semantic abstraction. Given initial and final world states, the task is to distinguish the proper action (\textsc{WorldPrediction-WM}) or the properly ordered sequence of actions (\textsc{WorldPrediction-PP}) from a set of counterfactual distractors. This discriminative task setup enable us to evaluate different types of world models and planners and realize a thorough comparison across different hypothesis. The benchmark represents states and actions using visual observations. In order to prevent models from exploiting low-level continuity cues in background scenes, we provide “action equivalents” – identical actions observed in different contexts – as candidates for selection. This benchmark is grounded in a formal framework of partially observable semi-MDP, ensuring better reliability and robustness of the evaluation. We conduct extensive human filtering and validation on our benchmark and show that current frontier models barely achieve 57\% accuracy on \textsc{WorldPrediction-WM} and 38\% on \textsc{WorldPrediction-PP} whereas humans are able to solve both tasks perfectly.}

\date{\today}
\correspondence{Pascale Fung at \email{pascalefung@meta.com}}

\begin{document}

\maketitle

\section{Introduction}
\label{sec:Introduction}

Advanced machine intelligence relies critically on two foundational capabilities: world modeling and procedural planning~\citep{lecun2022path}. World modeling \citep{ha2018world} allows agents to internally simulate future world states, enabling them to optimize their actions accordingly without trial-and-error in the real world or relying exclusively on explicit reward signals. Procedural planning~\citep{chang2020procedure} involves strategically determining ordered sequences of actions to achieve long-horizon goals. These capabilities represent key steps toward developing AI agents that can reason effectively, act responsibly, and interact smartly with complex environments.

Recent advances in low-level world modeling and planning have achieved significant progress in intuitive physics understanding~\citep{garrido2025intuitive}, robotic motion control~\citep{zhou2024dino}, navigation~\citep{koh2021pathdreamer, bar2024navigation}, and autonomous driving~\citep{wang2024driving}. These scenarios typically involve precise physical dynamics and high-frequency control without any semantic or temporal abstraction. However, skilled human activities require reasoning at a higher level, where individual actions span longer, non-uniform durations and encapsulate multiple lower-level primitive actions~\citep{sutton1999between}. Existing benchmarks focus on narrow task-specific setups~\citep{wang2023event,valmeekam2023planbench} or over-constrain their benchmark to specific model architectures, e.g., only for video generation~\citep{duan2025worldscore}, or only for text-based planning~\citep{choi2024lota} in some instances.

We propose \textsc{WorldPrediction}, a benchmark for evaluating high-level world modeling and long-horizon procedural planning in very diverse domains. It consists of two sub-benchmarks: \textsc{WorldPrediction-WM} assesses whether the model understands the causalities of semantically and temporally abstract actions in real-world skilled human activities; \textsc{WorldPrediction-PP} further extends the evaluation to procedural planning over extended temporal horizons, in contrast to existing benchmarks that typically focus on short spans of only 3-4 steps~\citep{chang2020procedure}. Key features of the \textsc{WorldPrediction} benchmark include:

\begin{figure}
    \centering
    \includegraphics[width=0.66\linewidth]{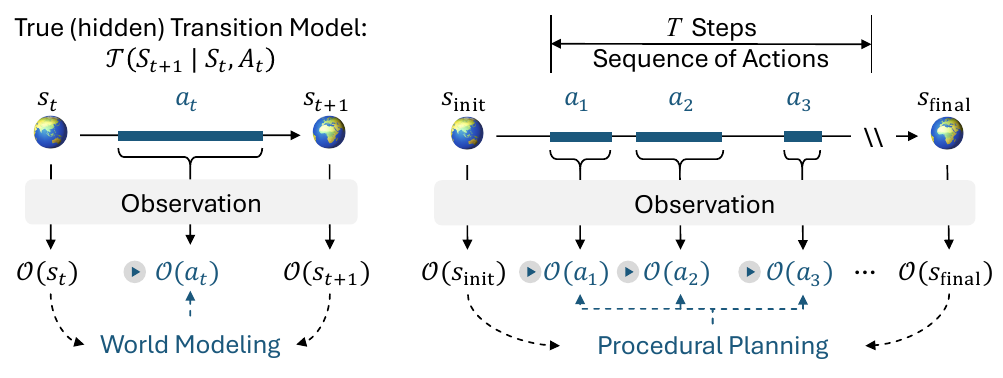}
    \caption{\textbf{Theoretical formulation of \textsc{WorldPrediction}}. Latent world states ($s$) and high-level actions ($a$) evolve according to a hidden transition model $\mathcal{T}$, which is not directly accessible. Instead, an observation model $\mathcal{O}$ maps these latent variables into visual observations, producing images $\mathcal{O}(s)$ depicting states and video segments $\mathcal{O}(a)$ depicting actions.  
    }
    \label{fig:formulation}
\end{figure}

\textbf{1) Diverse Actions and Tasks}.
The benchmark covers a broad spectrum of human activities, such as food preparation, household repair, technical maintenance, furniture assembly, health care, etc. Samples are sourced from five datasets – COIN~\citep{tang2019coin}, CrossTask~\citep{zhukov2019cross}, EgoExo4D~\citep{grauman2024ego}, EPIC-KITCHEN-100~\citep{damen2022epickitchens}, and IKEA-ASM~\citep{ben2021ikea} – encompassing instructional web videos as well as egocentric and exocentric recordings of skilled human activities. This extensive coverage ensures a holistic evaluation of model capabilities.

\textbf{2) Discriminative Formulation}.
The benchmark adopts a multiple-choice task formulation, where models select correct actions or action sequences from a set of counterfactual distractors. It facilitates direct comparisons between diverse world model/planner architectures (\textit{e.g.}, predictive vs. generative), and modality representations (\textit{e.g.}, VLMs vs. diffusion). Additionally, it accommodates the intrinsic variability in real-world activities – where multiple valid solutions exist for the same goal – by tasking the models to identify the most plausible ones rather than requiring the exact reproduction of one particular plan.

\textbf{3) Shortcut Mitigation}.
The benchmark represents states and actions using visual observations. To discourage models from exploiting superficial background continuity cues, we provide ``action equivalents’’: identical actions captured in varying backgrounds or observed from different viewpoints as the action candidates for selection. This strategy effectively reduces superfluous correlations between initial/final states and the ground-truth actions, ensuring the benchmark accurately evaluates the understanding of action-state causality and the true action sequencing capabilities.

The design of \textsc{WorldPrediction} is grounded in a mathematical framework inspired by the \textbf{Partially Observable Semi-Markov Decision Process} (POSMDP) \citep{kaelbling1998planning,silver2010monte}. This framework models partial observability inherent in images and videos, and captures the semantic and temporal abstraction characteristics of high-level actions \citep{sutton1999between}. The framework provides principles that systematically guide our data curation and sample validation processes.

After carefully curating the samples in \textsc{WorldPrediction}, we establish baseline performance on \textsc{WorldPrediction} using several state-of-the-art (SOTA) approaches, including vision-language models (VLMs), Socratic large language models (LLMs), video diffusion models, and Open-Event Procedural Planning (OEPP) models~\citep{wu2024open}. Overall results on \textsc{WorldPrediction} demonstrate that while better perception on larger models yields expected improvements, a substantial gap still remains between the highest-performing models (57.0\% on \textsc{WorldPrediction-WM} and 38.1\% on \textsc{WorldPrediction-PP}) and human performance, which achieves perfect results on both tasks

\section{Related Works}
\label{sec:Related Works}

\subsection{Evaluation of World Models}

World Modeling is a fundamental capability of autonomous intelligent systems \citep{lecun2022path}, which consists in leveraging an internal representation of the world to predict and understand how the state of the world evolves under different perturbations and actions. Recent efforts in world modeling generally fall into two broad categories: \emph{predictive} models, which predicts future latent representations of the world \citep{assran2023self,bardesrevisiting,zhou2024dino}, and \emph{generative} models, which simulate future states directly in observation space \citep{yang2023learning,bruce2024genie,nvidia2025cosmosworldfoundationmodel}. In practice, due to the complexity of the real world, existing world models have been adopted either in synthetic environments \citep{kim2023imagine,hafner2023mastering,garrido2024learning,gupta2024context}, or in real world environments with relatively constrained action spaces such as low-level \emph{robotics} \citep{hafnerdream,wu2023daydreamer,mendoncastructured,zhou2024robodreamer} with manipulation-based actions, \emph{autonomous driving} \citep{hu2023gaia,guan2024world,wang2024drivedreamer,wang2024driving} with vehicle control actions, and \emph{navigation} \citep{koh2021pathdreamer,shah2023lm,bar2024navigation} with spatial movement actions. 

There is currently no unified standard for evaluating world modeling. Existing benchmarks are often limited in scope, focusing on narrow, task-specific setups~\citep{wang2023bytesized32}, or are tightly coupled to architectural assumptions, which limits their general applicability. Some methods adopt a Visual Question Answering (VQA)-style evaluation~\citep{he2024mmworld}, requiring models to produce textual outputs that evaluate expert knowledge through visual understanding. Others focus exclusively on the quality of generated scenes, an approach that primarily suits video generation models~\citep{li2025worldmodelbench,duan2025worldscore}. For large language models (LLMs), current benchmarks either evaluate world generation through text~\citep{hu2025text2world} or assess decision-making within text-described scenarios~\citep{yang2024evaluating}.
In contrast, our proposed benchmark is designed to be both architecture-agnostic and task-agnostic, accommodating a wide variety of world model formulations. Importantly, \textsc{WorldPrediction} is the first to emphasize human-centric activities—going beyond simple object state transitions~\citep{xue2024learning} to evaluate a model’s understanding of dynamic human behaviors in complex environments.

\subsection{Evaluation of Procedural Planning}

Given an initial and final state at a longer horizon, Procedural Planning refers to the ability of predicting a sequence of actions which would bring the initial state towards the final state. While that formulation is especially present in robotic control \citep{sun2022plate,lynch2023interactive} for low-level manipulation tasks, in this work we focus on human-centered procedural planning with higher-level actions (e.g., ``remove the battery'', ``attach a table leg'') \citep{ben2021ikea,damen2022epickitchens}, mostly from instructional videos \citep{chang2020procedure,tang2019coin,zhukov2019cross}, which inherently involves deeper semantic reasoning and abstraction of granular actions. In this context, most of the current approaches either try to learn the action space \citep{zhao2022p3iv,niuschema,li2023skip} or leverage LLMs \citep{liu023language,wang2023event,islam2024propose} to generate abstracted procedural steps as high-level procedural planning is mostly evaluated in a constrained window of 3 to 4 steps.

Recent benchmarks have attempted to broaden the scope of procedural planning by integrating simulated environments and language-based reasoning~\citep{li2024embodied,choi2024lota}, or by evaluating natural task sequences such as travel planning or household routines~\citep{valmeekam2023planbench,zheng2024natural}. Others incorporate explicit path-based planning to test logical consistency and feasibility~\citep{aghzal2024can}. Despite these efforts, most benchmarks remain narrowly scoped in terms of domains and are heavily focused on LLM-centric evaluations, using textual outputs as proxies for structured plans. While this reflects the interpretability of language in capturing abstract reasoning, such benchmarks often ignore perceptual grounding or rely on synthetic visual inputs. Recent works made attempts to expand the scope of procedural planning~\citep{wu2024open,patel2023pretrained}, as the evaluation of the task is still over-reliant on human-annotated text labels of actions to convey interpretable plans, which motivates the formulation of our label-free procedural planning evaluation in \textsc{WorldPrediction}.

\section{The WorldPrediction Benchmark}
\label{sec:WorldPrediction}

\subsection{Theoretical Formulation}

We begin by formally defining a mathematical framework that provides the foundation for building the \textsc{WorldPrediction} benchmark. This formulation integrates elements from Partially Observable MDPs~\citep{kaelbling1998planning} and Semi-MDPs~\citep{sutton1999between} to accurately capture the complex dynamics inherent in human activity videos. Formally, we represent this framework as a tuple $\langle\mathcal{S, A, T, O}\rangle$:

\textbf{World States} $s \in \mathcal{S}$ constitute the continuous latent space representing the full underlying configuration of the environment. These states, although comprehensive, cannot be directly accessed and must instead be inferred from partial visual observations. Crucially, not all elements of a state are equally relevant to a given task: we distinguish between \textbf{task-relevant} components, which directly affect the causal outcomes of actions and are essential for achieving goals, and \textbf{task-irrelevant} components, representing background details or contextual information that do not influence the task.

\textbf{(High-level) Actions} $\mathcal{A} = {A_{1}, A_2, \dots, A_N}$ represent the vocabulary of all possible actions. Here, high-level'' is characterized by both \textbf{semantic and temporal abstraction}, differentiating them from low-level continuous controls executed at fixed intervals. Each high-level action encapsulates several lower-level motor primitives or sub-actions. This can be modeled by \textit{options} in Semi-MDPs, which are defined by a policy over low-level primitives, a termination condition, and a set of world states that allow that specific action. All components are dependent on the current environmental states, ensuring adaptation to varying contexts, as illustrated in Fig.~\ref{fig:action_equivalents}. To distinguish from abstract action categories, we use the notation $a \in A$ to represent an action instance performed in a specific context $s$ (\textit{e.g.,} $A_i$ represents cut potato’’ and $a \in A_i$ is the muscle motion sequence of cutting a potato in one particular kitchen setting).

\begin{figure*}[!t]
    \centering
    \includegraphics[width=\textwidth, trim=0 110 50 0, clip]{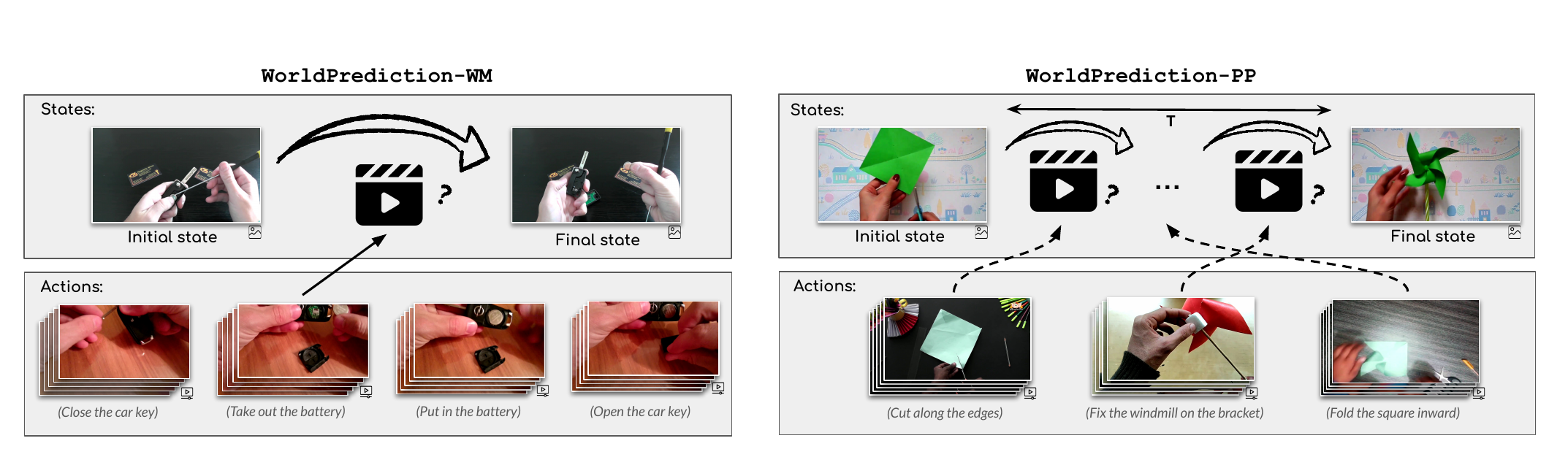}
    \vspace{-15pt}
    \caption{\textsc{WorldPrediction-WM} and \textsc{WorldPrediction-PP} task formulation. For World Modeling, the objective is to select which action clip depicts the transition from initial to final state. For Procedural Planning, the objective is to select which sequence of action clips ($T \in [3, 10]$) is correctly ordered to depict the transition from initial to final state. The actual samples do not contain any text, here the actions are annotated for visualization purposes.}
    \label{fig:worldprediction_main}
    \vspace{-5pt}
\end{figure*}

\textbf{Transition Model} $\mathcal{T}$ specifies the true underlying mechanism governing how world states evolve over time – after an action $ a_t$ is taken at $s_t$, the world state transitions to a new state $s_{t+1}$ with a probability of $\mathcal{T}\bigl(s_{t+1}\mid s_t, a_t\bigr)$. In real-world, non-simulated environments, this transition mechanism is hidden and thus inaccessible; agents must approximate it by learning a \textbf{world model}. It enables reasoning and planning without relying directly on explicit reward signals or costly trial-and-error interactions in the real world.

\textbf{Observation Model} $\mathcal{O}$ maps latent world states or performed actions to corresponding sensory signals, \textit{i.e.,} an image $\mathcal{O}(s_t)$ and a video segment $\mathcal{O}(a_t)$. Due to intrinsic limitations of perception devices (\textit{e.g.,} occlusions, resolution, or viewpoint constraints), they only provide imperfect views of the underlying true state or the performed action, and also contain an excessive amount of task-irrelevant background information. To address these challenges brought by \textbf{partial observability}, our benchmark incorporates two strategies detailed in \S\ref{sec:benchmark_design}: \textit{observability filtering}, which excludes samples lacking sufficient visual evidence of action outcomes, and \textit{action equivalents}, which mitigate the shortcut based on superficial background continuity cues.

Given the tuple $\langle\mathcal{S, A, T, O}\rangle$, we can formally characterize the underlying data-generative process of human activity videos as follows. Beginning from an initial latent state $s_0$, a human agent decides to perform an action $a_0 \in A_i$. The transition model $\mathcal{T}$ subsequently generates the next latent state $s_1$ conditioned on $s_0$ and $a_0$. This process iterates over multiple steps. Through the observation model $\mathcal{O}$, each latent state $s_t$ and action $a_t$ is mapped to visual observations, yielding the observed video sequence: $[\mathcal{O}(s_0), \mathcal{O}(a_0), \mathcal{O}(s_1), \mathcal{O}(a_1), \dots, \mathcal{O}(s_T)]$.

\subsection{Benchmark Objectives}

Our primary goal is to measure a model’s ability to understand real-world state transitions and the causal factors that drive them. Concretely, we focus on capturing how an initial world configuration evolves into a new configuration when subjected to a particular high-level action. This predictive ability, known as \emph{world modeling}, is formalized by having a learned function $\mathcal{W}$ approximate the true underlying transition model $\mathcal{T}$. Under a suitable divergence metric $\mathcal{D}$, the performance of a world model can naturally be defined as:

\begin{equation}
\mathcal{D}\left(\mathcal{W}(s_{t+1}\mid s_t, a_t) \,\|\, \mathcal{T}(s_{t+1}\mid s_t, a_t)\right).
\end{equation}

\begin{figure}
    \centering
    \includegraphics[width=0.75\linewidth]{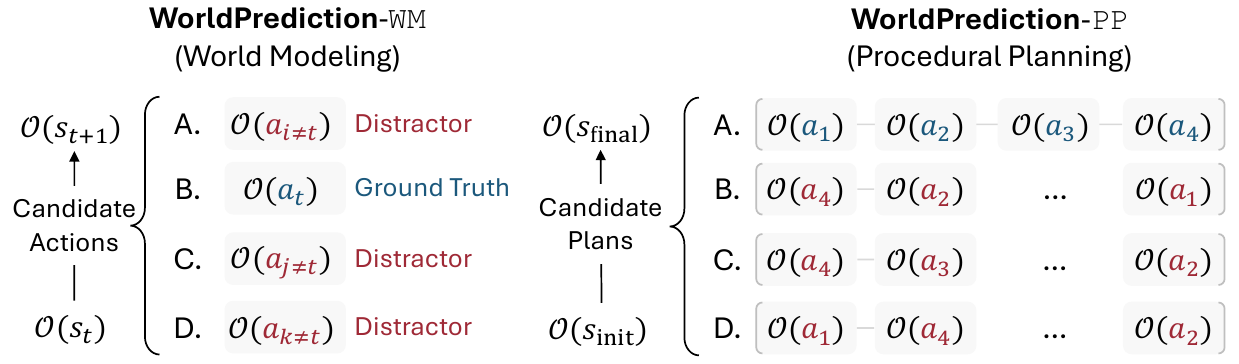}
    \caption{\textbf{Discriminative task formulation of \textsc{WorldPrediction}}. 
    Each sample includes a pair of visual observation of states along with a set of candidate actions or action sequences. Models must identify the correct one responsible for the observed state transition among distractors. Note that every $\mathcal{O}(a)$ is substituted by its action equivalent to avoid trivial background continuity shortcut.}
    \vspace{-5pt}
    \label{fig:task_design}
\end{figure}

Intuitively, a high-performing world model assigns a higher likelihood to correct state transitions $(s_t, a_t)\rightarrow s_{t+1}$ and a lower likelihood to incorrect transitions involving counterfactual combinations of states and actions. Formally, given a learned transition model $\mathcal{W}$, this implies the inequality $\mathcal{W}(s_{t+1}\mid s_t, a_t)  > \mathcal{W}(s_{t+1}\mid s_t, a_{j})$ for any counterfactual action $a_j \neq a_t$. Because we specifically focus on evaluating the understanding of \textit{high-level} actions rather than low-level primitives, we define this criterion at the action-category level: given the true action category $A^*$ corresponding to the correct action $a_t$, we empirically approximate the theoretical divergence by verifying whether the model assigns the highest likelihood to the correct action category responsible for the observed transition:

\begin{equation}
A^* \stackrel{?}{=} \arg\max_{A \in \mathcal{A}} \mathcal{W}(s_{t+1}\mid s_t, A).
\label{eq:objective-wm}
\end{equation}

This formulation probes a model’s approximation of the hidden transition model $\mathcal{T}$ by evaluating how well the causal relationship between $(s_{t}, a)$ and $s_{t+1}$ is captured. To have a robust approximation of $\mathcal{T}$, world models should learn to capture and discriminate the various ways in which actions transform the latent world state, rather than simply matching superficial or spurious correlations between states and actions, which we ensure in our design detailed later in section \ref{sec:benchmark_design}.

This argmax formulation of evaluation also enables a natural extension to multi-step procedural planning evaluation, where a \textit{plan} consisting of a sequence of actions can be viewed as a single \emph{``macro-action’’}, linking distant initial and final states. Specifically, given an initial state $s_{\text{init}}$ and a final state $s_{\text{final}}$ separated by $T$ high-level actions, the objective is to select the correct \textit{ordered} sequence of actions $\mathcal{P}^*=(a_1,\ldots,a_T)$ responsible for this long-horizon transition:

\begin{equation}
\mathcal{P}^*
\stackrel{?}{=}
\mathop{\arg\max}\limits_{\mathcal{P}\,\in\, \mathcal{A}^T}
 \mathcal{W} \bigl( s_{\text{final}}\mid s_{\text{init}},\, \mathcal{P}\bigr),
 \label{eq:objective-pp}
\end{equation}

where $\hat{\mathcal{P}} = (\hat{a}1, \dots, \hat{a}T)$ denotes the correct action sequence that transitions $s{\text{init}}$ to $s{\text{final}}$, and $\mathcal{A}^T$ denotes candidate plans of all possible arrangements of $T$-step action sequences. In principle, if all intermediate states $(s_{2}, \dots, s_{T-1})$ were known, solving procedural planning would reduce to solving $T$ successive world modeling steps. However, since these intermediate states are unobserved, the model must internally infer them, effectively reasoning about the entire multi-step causal chain.

\subsection{Benchmark Design}
\label{sec:benchmark_design}

\textbf{Task Formulation.} We now outline the design of our benchmark. As the true underlying states and transitions in real-world scenarios are not directly accessible, our benchmark instead leverages \emph{visual observations}—images or video clips—as cues to infer the true states and actions. We present \textsc{WorldPrediction-WM} and \textsc{WorldPrediction-PP}, two benchmarks respectively evaluating world modeling (Eq.\ref{eq:objective-wm}) and procedural planning (Eq.\ref{eq:objective-pp}) capabilities as shown in Figure \ref{fig:worldprediction_main}. Concretely, each sample consists of:

\begin{itemize}
    \item \textbf{State Observations}: Static images capturing the environment's configuration before and after the action(s) being taken, denoted as $\mathcal{O}(s_{t})$, $\mathcal{O}(s_{t+1})$ for \textsc{WorldPrediction-WM} and $\mathcal{O}(s_{\text{init}})$,  $\mathcal{O}(s_{\text{final}})$ for \textsc{WorldPrediction-PP}.
    
    \item \textbf{Action / Plan Candidates}: The search space of the argmax operation in Eq.~\ref{eq:objective-wm}  and Eq.~\ref{eq:objective-pp}, containing one ground truth ($A^*$ or $\mathcal{P}^*$) and several distractors. To enhance computational efficiency, the candidate pool can be limited to a small subset of the complete action space $\mathcal{A}$ or plan space $\mathcal{A}^T$.

\end{itemize}

\begin{figure}
    \centering
    \includegraphics[width=0.75\linewidth]{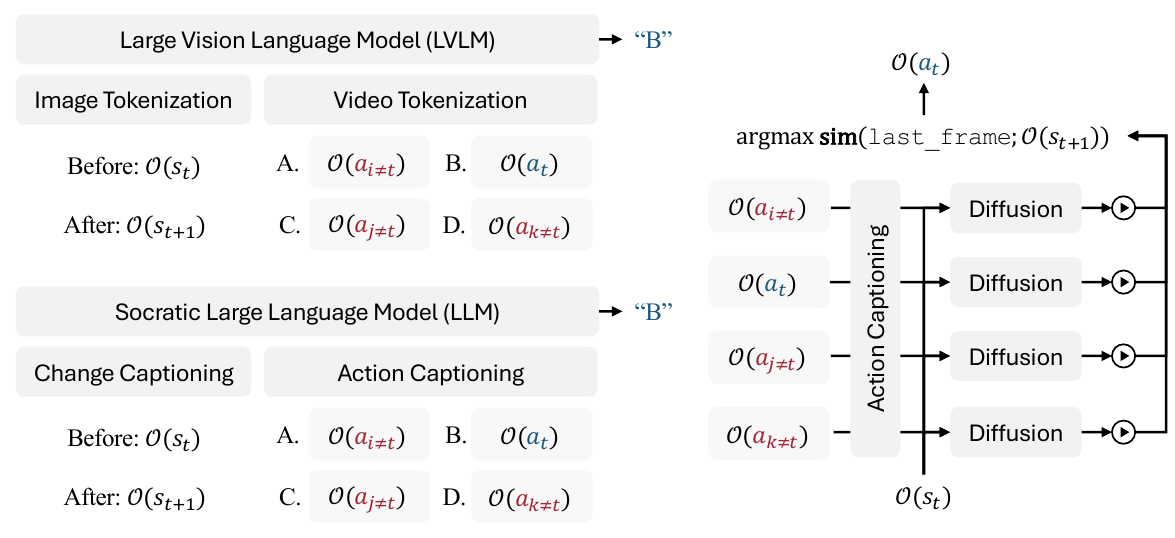}
    \caption{\textbf{Baseline models}. 
    VLMs directly encode visual observations, while Socratic LLMs first generate textual captions describing state changes and candidate actions, then select the action through text-only reasoning. Video diffusion models generate future observations conditioned on action captions, selecting the action by comparing final generated frame and the desired $\mathcal{O}(s_{t+1})$.}
    \label{fig:world_models}
    \vspace{-5pt}
\end{figure}

Models must select which action (or action sequence) accounts for the observed change in $\mathcal{O}(s_{t}) \rightarrow \mathcal{O}(s_{t+1})$ or $\mathcal{O}(s_{\text{init}}) \rightarrow \mathcal{O}(s_{\text{final}})$, providing a clear evaluation of world modeling and procedural planning. This discriminative multiple-choice setup (illustrated in Fig.\ref{fig:task_design}) directly aligns with our theoretical grounding (Eq.\ref{eq:objective-wm} and Eq.~\ref{eq:objective-pp}) and also offers several practical advantages. It universally accommodates different types of world models and planners (\textit{e.g.,} models using different architectures, generating different modalities to represent the predicted states). Additionally, by using only raw visual observations, we remove the reliance on human-annotated text labels as done in previous benchmarks \citep{chang2020procedure}, ensuring an unbiased evaluation\footnote{Although models can still generate captions from visual observations (as in Socratic LLM baselines provided in \S\ref{sec:baselines}), we view them as models’ internal perceptual representations.}.

\textbf{Action Equivalents.} Due to being purely observation-based, an important challenge in the construction of our benchmark is to prevent models from exploiting trivial continuity cues to identify the correct action or sequence. Specifically, if the same camera viewpoint, background objects, or other task-irrelevant visual elements are preserved across the state observations as well as the ground-truth action segment, then a model might simply match low-level features without learning the true causal relationship between action content and state transitions. Such an approach would result in models failing to capture the \emph{semantic and temporal abstractions} of high-level actions. To mitigate this shortcut, we employ \textbf{action equivalents} (shown in Appendix, Fig. \ref{fig:action_equivalents}). For each high-level action category $A_{i}$, there exists a set of observations which depict it being performed in visually different environments or from a significantly different viewpoint (\textit{e.g.,} egocentric vs exocentric). Concretely, we use that set to replace the ground-truth observation action with one of its action equivalents and re-sample distractors from the same environment of the action equivalent for \textsc{WorldPrediction-WM}, and re-shuffle the new sequence of equivalent actions for \textsc{WorldPrediction-PP}.

\begin{figure*}[t]
    \centering
    \includegraphics[width=\linewidth]{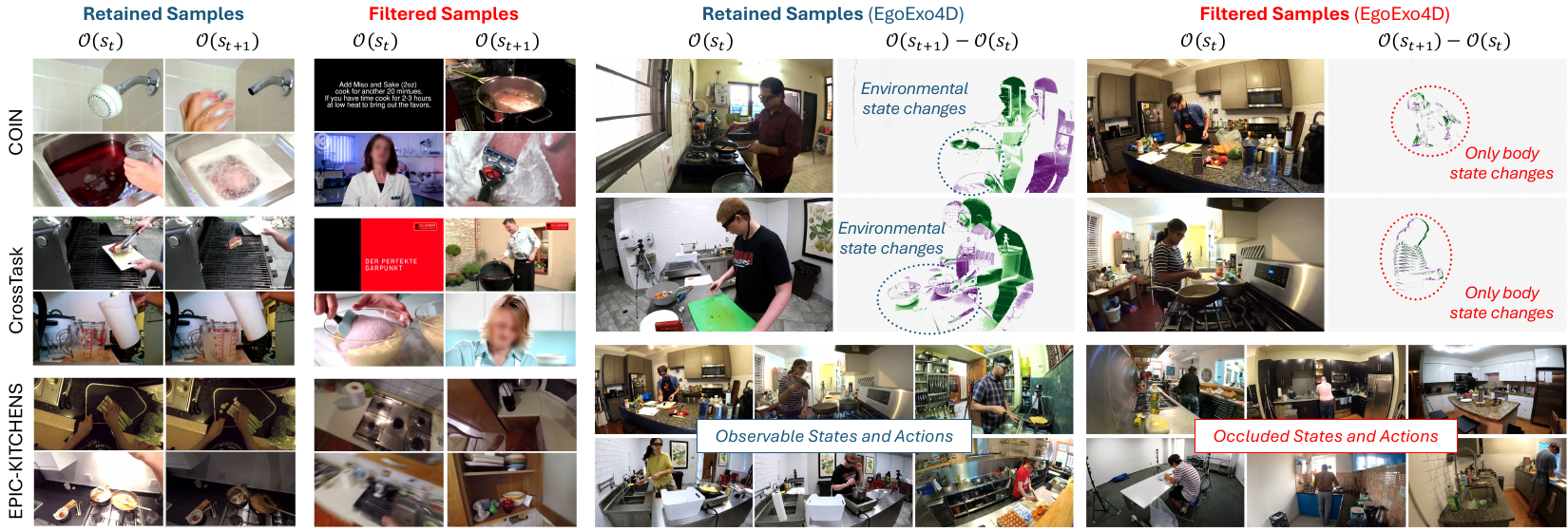}
    \caption{\textbf{Sample Filtering in \textsc{WorldPrediction}.} 
    Samples are retained only if state observations clearly show meaningful environmental changes resulting from actions. Samples are filtered out if they exhibit excessive viewpoint shifts, only contain minor body movements without clear environmental changes, or severe occlusions, which all makes causal inference challenging.}
    \label{fig:filterng}
    \vspace{-5pt}
\end{figure*}

\textbf{Observability Filtering.} Under the \emph{partial observability} assumption, task-relevant elements of the environment can sometimes fail to be captured in state observations. When the evidence needed to infer what changed—and thus which action caused the transition—is missing, the ambiguity increases significantly and the task becomes nearly impossible even for humans. There are two main causes for failing to capture the action-relevant state observation: \textbf{noisy observation} due to video edits or drastic camera field-of-view shifts, and \textbf{occlusions} due to different entities blocking the view of task-relevant objects.

To remove samples with noisy observation, we employ the assumption that noisy observation usually causes larger changes in semantic feature space. Specifically, we compute the distance $d$ between the visual features for both state observations: $d = |\phi(\mathcal{O}(s_{\text{init}})) - \phi(\mathcal{O}(s_{\text{final}}))|2$ using a pretrained vision encoder $\phi(\cdot)$, and we only keep pairs $(\mathcal{O}(s{\text{init}}), \mathcal{O}(s_{\text{final}}))$ whose similarity score is smaller than a certain threshold, thus removing samples where the scene changes so drastically that no coherent causal link can be reliably inferred. The left side of Fig. \ref{fig:filterng} provides an example of this filtering. This filtering process can be seen as a coarse classifier that eliminates a large portion of the bad state observations by relying on the assumption that observations which are too different are highly likely to miss task-relevant information in at least one of the two states. This assumption also aligns with the POMDP formulation: consecutive observations of the same environment should not appear uncorrelated if they reflect smoothly evolving states in the real world.

Additionally, we filter out exocentric state observations where the human performing the action has their back turned toward the camera (or otherwise heavily obstructing the view, as shown in the bottom right of Fig. \ref{fig:filterng}), as in such cases it becomes exceedingly difficult to discern the critical objects or interactions relevant to the action. Consequently, the remaining samples more consistently capture the essential task-relevant cues for modeling and evaluating high-level transitions, aligning with the \textbf{partial observability} principle in a controlled yet realistic setting.

\subsection{Benchmark Implementation}
\label{sec:benchmark_implementation}

\textbf{Dataset Sources.} \textsc{WorldPrediction} incorporates five publicly available datasets to ensure broad coverage and representativity of skilled human activities:
\begin{itemize}
\item \textbf{COIN}\citep{tang2019coin}: provides instructional web videos covering diverse procedural tasks, such as cooking and household repairs.
\item \textbf{CrossTask}\citep{zhukov2019cross}: consists of instructional web videos capturing diverse everyday activities.
\item \textbf{EgoExo4D}\citep{grauman2024ego}: provides temporally-aligned egocentric and multi-view exocentric videos. We focus specifically on the \texttt{cooking} and \texttt{healthcare} subsets, which emphasize procedural human activities.
\item \textbf{EPIC-KITCHENS-100}\citep{damen2022epickitchens}: is a large-scale egocentric dataset of kitchen tasks with detailed annotations, capturing fine-grained interactions.
\item \textbf{IKEA-ASM}~\citep{ben2021ikea}: features clear exocentric instructional videos of furniture assembly, providing structured action sequences in controlled environments.
\end{itemize}

We use official dataset splits for evaluation: the test split for COIN and validation splits for CrossTask, EPIC-KITCHENS-100, EgoExo4D, and IKEA-ASM. For \textsc{WorldPrediction-PP}, we use a number of action steps $T \in {3, 4}$ for COIN and CrossTask, and $T\in{3,4, \dots, 10}$ for the remaining. The action sequences are sampled in a sliding window fashion following previous works. The statistics for the \textsc{WorldPrediction} benchmark dataset are detailed in Table \ref{tab:dataset_statistics}, with additional information provided in Appendix \ref{app:additional_dataset_info}.

\textbf{Distractor Sampling.} To rigorously test action discrimination, each correct action is presented alongside three distractors, resulting in four total candidates per sample. For \textsc{WorldPrediction-WM}, distractors are plausible alternative actions drawn from the same task context (\textit{i.e.,} same video) but incompatible with the observed state transition. For \textsc{WorldPrediction-PP}, distractors are generated by shuffling the ground-truth action sequences, preserving action-level plausibility while disrupting temporal correctness.

\begin{table}[t]
  \centering
  \resizebox{\linewidth}{!}{%
    \begin{tabular}{lcccccc}
      \toprule
      \multirow{2}{*}{\textbf{Dataset}} &
        \multicolumn{3}{c}{\textsc{WorldPrediction-WM}} &
        \multicolumn{3}{c}{\textsc{WorldPrediction-PP}} \\
      \cmidrule(lr){2-4} \cmidrule(lr){5-7}
       & \textbf{\# Samples} & \textbf{\# Unique Actions} & \textbf{Avg.\ Duration (s)} &
         \textbf{\# Samples} & \textbf{\# Unique Actions} & \textbf{Avg.\ Duration (s)} \\
      \midrule
      COIN                  & 236 & 532  & 13.16 & 243 & 285 & 14.70 \\
      CrossTask             & 109 & 194  &  9.17 &  58 &  65 &  7.53 \\
      IKEA ASM              & 159 & 185  &  9.02 & 136 &  43 &  6.48 \\
      EgoExo4D              & 128 & 128  & 11.71 &  76 & 180 & 11.23 \\
      EPIC-KITCHENS-100     & 193 & 561  &  6.25 &  57 & 176 &  3.47 \\
      \midrule
      \textbf{\textsc{WorldPrediction} (All)}          & 825 & 1800 & 10.02 & 570 & 749 &  9.38 \\
      \bottomrule
    \end{tabular}%
  }
  \caption{\textsc{WorldPrediction} dataset statistics (number of samples, actions, and average action duration) for both tasks}
  \label{tab:dataset_statistics}
\end{table}

\textbf{Action Equivalent Retrieval.} To mitigate shortcut learning from low-level visual continuity cues, we employ \textit{action equivalents}: visually different yet semantically identical actions captured in alternate backgrounds or viewpoints, as detailed in \S\ref{sec:benchmark_design}. For COIN, CrossTask, EPIC-KITCHENS-100, and IKEA-ASM, actions sharing the same textual label constitute equivalents. For EgoExo4D, where explicit temporal boundaries are unavailable, we segment actions by computing midpoints between consecutive timestamps and discard segments shorter than 5 seconds. We select the egocentric view for actions to clearly observe detailed hand movements and use exocentric viewpoints for state observations due to their comprehensive scene coverage.

\textbf{Sample Filtering.} To filter out noisy observations, we compute distances between visual features of initial and final states using pretrained visual embeddings (DINOv2 \citep{oquab2024dinov2}). Samples exceeding predefined thresholds (2.75 for \textsc{WorldPrediction-WM}, 10 for \textsc{WorldPrediction-PP}) are excluded due to excessively drastic or incoherent scene transitions. For EgoExo4D, we additionally remove samples in which critical task-relevant visual information is obstructed by the human subject. This is implemented by prompting a VLM with \texttt{“Is the main person not showing their back and what they are doing with hands being clearly visible?”}. We further remove samples where there is too little difference between their initial and final states. These samples usually correspond to a static segment in an instructional video, or only slight body movement in EgoExo4D videos (as shown in Fig.~\ref{fig:filterng}). IKEA-ASM features clear and comprehensive observations, requiring no additional filtering.

\section{Evaluation Results}
\label{sec:baselines}

\subsection{Models}

We establish initial baseline performance on \textsc{WorldPrediction} using VLMs, Socratic LLMs, and video diffusion models, and Open-Event Procedural Planning (OEPP) models. Among them, VLMs and Socratic LLMs serve as both world models and procedural planners due to their flexibility, while diffusion is tailored to world modeling and OEPP is only for planning. These baselines are chosen for their popularity and straightforward implementation, serving primarily to provide initial reference points for future research.

\begin{table}[t]
  \centering
  \begin{minipage}[t]{0.48\textwidth}
    \vspace*{10pt}
    \centering
    \resizebox{\linewidth}{!}{%
        \begin{tabular}{cccc}
          \toprule
          \multicolumn{2}{c}{\textbf{World Model / Planner}} &
            \textbf{\begin{tabular}[c]{@{}c@{}}WorldPrediction\\ -\texttt{WM}\end{tabular}} &
            \textbf{\begin{tabular}[c]{@{}c@{}}WorldPrediction\\ -\texttt{PP}\end{tabular}} \\ 
          \midrule
    
           & InternVL2.5 (2B)       & 20.0  & 21.05 \\
           & InternVL2.5 (4B)       & 29.8  & 27.9 \\
           & InternVL2.5 (26B)      & 30.2  & 30.0 \\
           & InternVL2.5 (38B)      & 50.3  & 31.1 \\
          \cmidrule(r){2-4}
           & Qwen2.5-VL (3B)        & 21.6  & 29.1 \\
           & Qwen2.5-VL (7B)        & 45.5  & 32.5 \\
           & Qwen2.5-VL (32B)       & 49.0  & 33.5 \\
           \multirow{-8.4}{*}{\begin{tabular}[c]{@{}c@{}}\textbf{VLMs}\end{tabular}}
           & Qwen2.5-VL (72B)       & \textbf{57.0}  & \textbf{36.7} \\
    
          \midrule
    
           & Llama-3.1 (8B)         & 48.7  & 26.7 \\
           & Llama-3.1 (70B)        & 49.8  & 31.2 \\
           & Llama-3.3 (70B)        & 52.2  & 35.1 \\
           & Llama-4 Scout          & 52.7  & 32.8 \\
           & Llama-4 Maverick       & 53.6  & 34.7 \\
          \cmidrule(r){2-4}
           & Qwen2.5 (3B)           & 44.0  & 25.6 \\
           & Qwen2.5 (7B)           & 49.1  & 28.4 \\
           & Qwen2.5 (32B)          & 39.2  & 29.1 \\
           & Qwen2.5 (72B)          & 48.5  & 30.7 \\
          \cmidrule(r){2-4}
           & DeepSeek-R1 (distilled)& 50.8 & 28.4   \\
          \cmidrule(r){2-4}
           & Gemini-2.0             & \textbf{55.6}  & 33.5 \\
           & GPT-4o                 & 52.0  & 33.7 \\
          \multirow{-14}{*}{\begin{tabular}[c]{@{}c@{}}\textbf{Socratic}\\\textbf{LLMs}\end{tabular}}
           & Claude-3.5-sonnet      & 53.3  & \textbf{38.1} \\ 
          \midrule
    
           & I2VGenXL               & 26.1  &      \\
           & I2VGenXL + DINOv2      & 26.7  &      \\
           & CogVideoX              & 30.1  &      \\
          \multirow{-4}{*}{\begin{tabular}[c]{@{}c@{}}\textbf{Video}\\\textbf{Diffusion}\end{tabular}}
           & CogVideoX + DINOv2     & 30.5  & \multirow{-4}{*}{N/A} \\ 
          \midrule
    
           & MLP                    & \multirow{3}{*}{N/A}   & 36.8 \\
           & Transformer            &       & 34.2 \\
          \multirow{-3}{*}{\textbf{OEPP}}
           & PDPP                   &       & 34.4 \\ 
          \bottomrule
        \end{tabular}%
        }
    \caption{Performance results on \textsc{\textsc{WorldPrediction-WM}} and \textsc{\textsc{WorldPrediction-PP}} w/ accuracy (\%).}
    \label{tab:performance_comparison}
  \end{minipage}%
  \hfill
  \begin{minipage}[t]{0.48\textwidth}
    \vspace*{0pt}
    \centering

    \includegraphics[width=\linewidth]{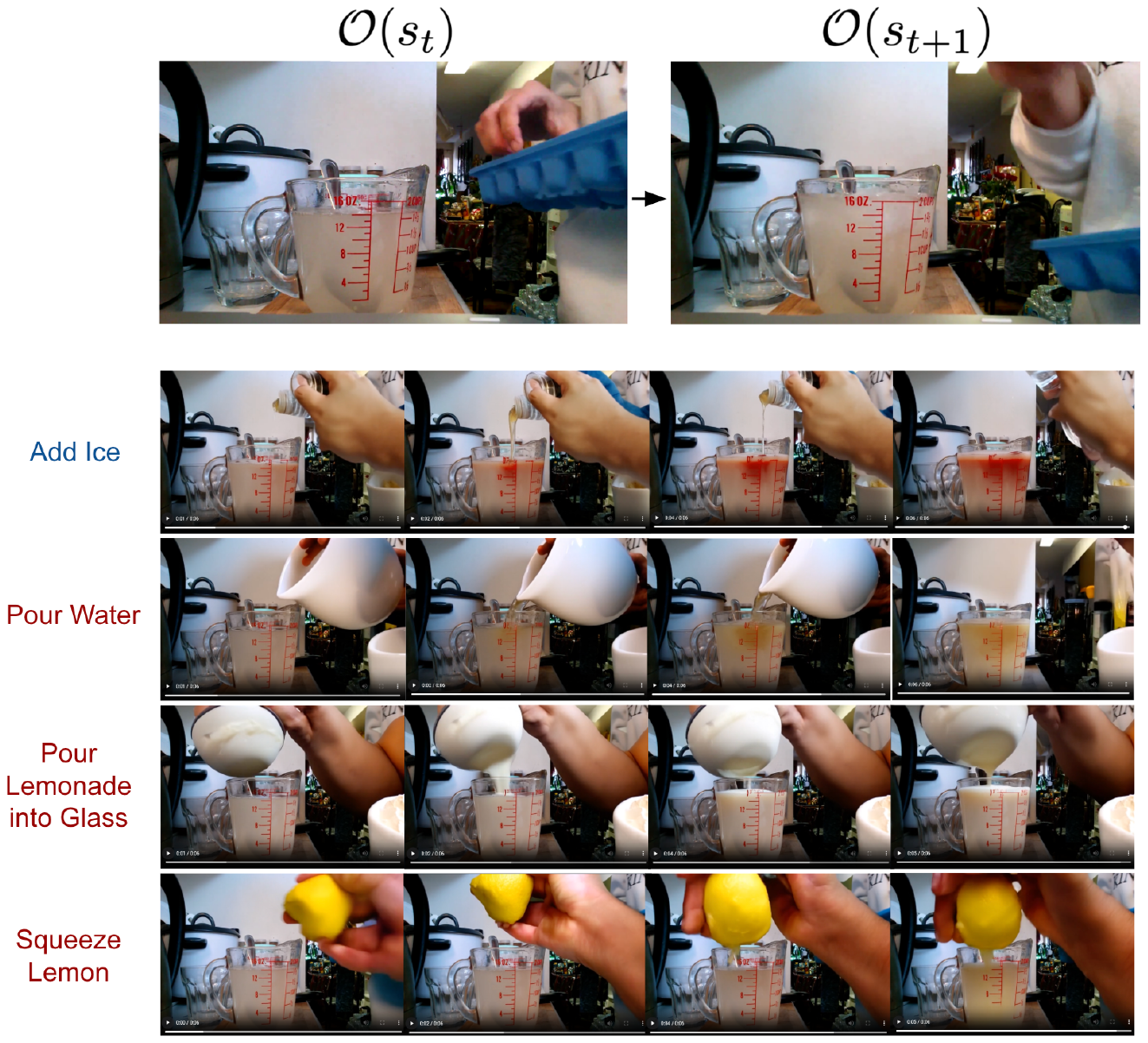}
    \captionof{figure}{Generated sample using CogVideoX-I2V}
    \label{fig:video_diffusion_generation}

    \vspace{1em}
  
    \resizebox{\linewidth}{!}{%
      \begin{tabular}{ccccc}
        \toprule
        \multicolumn{2}{c}{\textbf{Procedural Planner}} &
          \textbf{\begin{tabular}[c]{@{}c@{}}COIN,\\ CrossTask\end{tabular}} &
          \textbf{\begin{tabular}[c]{@{}c@{}}EgoExo4D, E‑100\\ IKEA‑ASM\end{tabular}} &
          \textbf{Overall} \\ \midrule
        VLMs &
          \begin{tabular}[c]{@{}c@{}}Qwen2.5VL\\ (72B)\end{tabular} &
          37.6 & 35.0 & 36.1 \\ \midrule
        \begin{tabular}[c]{@{}c@{}}Socratic\\ LLMs\end{tabular} &
          \begin{tabular}[c]{@{}c@{}}Llama‑3.3\\ (70B)\end{tabular} &
          34.3 & 41.0 & 37.4 \\ \midrule
         & MLP         & 42.3 & 26.5 & 36.8 \\
         & Transformer & 48.3 & 29.5 & 34.2 \\
        \multirow{-3}{*}{OEPP} & PDPP & 49.2 & 29.4 & 34.4 \\ \bottomrule
      \end{tabular}%
      }
    \captionof{table}{Detailed performance comparison of \textsc{WorldPrediction-PP}.}
    \label{tab:planning_ablation}

  \end{minipage}
\end{table}

\paragraph{VLMs.}
We use two state-of-the-art open-source VLM families: \texttt{Qwen2.5-VL} \citep{yang2024qwen2} and \texttt{InternVL2.5} \citep{chen2024internvl}. As shown in Fig.~\ref{fig:world_models}, to perform the \textsc{WorldPrediction} multiple-choice task, models are prompted with a structured multimodal query comprising images depicting the initial and final world states, video segments representing the candidate actions, along with textual instructions explaining the task and specifying the desired output format. We frame the task explicitly by instructing the model to select the most plausible action or the sequence of actions that cause the observed state transition.

\paragraph{Socratic LLMs.}
We evaluate the performance of Socratic LLMs~\citep{zeng2022socratic}, which decouple perception and reasoning into two distinct stages. Visual inputs are translated into textual descriptions through a VLM, then a text-only instruct-tuned LLM is prompted with these captions along with instructions, including structured task explanations and candidates. The LLM then employs textual reasoning to identify the action or sequence of actions most plausibly causing the observed state transitions. To obtain the textual descriptions, we utilized Qwen 2.5-VL (72B). For the text-only LLM, we evaluated five different LLM families with varying sizes, including Llama 3.1-Instruct (8B, 70B, 405B), Qwen 2.5-Instruct (3B, 7B, 14B, 72B), DeepSeekR1 (distilled version of Qwen-32B), GPT-4o, and Claude-3.5-Sonnet.

\paragraph{Video Diffusion Models.}
To assess generative world modeling capabilities, we also evaluate two image-conditioned video diffusion models: I2VGenXL \citep{zhang2023i2vgen} and CogVideoX-I2V \citep{hong2022cogvideo}, which directly generate the future state in pixel space. For inference, we provide the initial state observation $\mathcal{O}(s_t)$ as the grounding image and perform action captioning using a VLM to get a text description of each action candidate. The generated video is a visual representation of the state transition toward the final state observation $\mathcal{O}(s_{t+1})$. We select the most likely action candidate by identifying the generated segment whose last frame exhibits the smallest pixel-wise distance to $\mathcal{O}(s_{t+1})$.

\paragraph{OEPP Models.}
We reimplement OEPP models~\citep{wu2024open} and incorporate them into the \textsc{WorldPrediction-PP} task. OEPP performs planning using VideoCLIP~\citep{xu-etal-2021-videoclip} embeddings. Given initial and final observations, a planning model (either MLP, Transformer~\citep{vaswani2017attention}, or PDPP~\citep{wang2023pdpp}) is trained to generate $T$ text embeddings corresponding to a sequence of $T$ predicted actions. We embed all candidate plans into the same text embedding space and select the candidate that minimizes the distance with the generated embeddings.

\subsection{Performance Comparison}

Table~\ref{tab:performance_comparison} summarizes model performances on the \textsc{WorldPrediction} benchmark. In the \textsc{WorldPrediction-WM} task, smaller-scale VLMs perform near random chance levels, with InternVL2.5 (4B) and Qwen2.5-VL (3B) models notably struggling to produce outputs that choose from given options, resulting in 25\% and 77\% unparsable responses in WM and PP, respectively. There is a significant breakthrough in world modeling performance past a certain model scale, with a jump of roughly 20\% from 26B to 38B for InternVL2.5, and from 3B to 7B for Qwen2.5-VL. However, it is interesting to note that long-horizon procedural planning does not show a significant boost in performance with model size. Socratic LLMs, using high-quality captions generated by Qwen2.5-VL (72B), achieve comparable results to VLMs. The best-performing LLMs are the closed-source Gemini-2.0 for world modeling at 55.6\% and Claude-3.5 for procedural planning at 38.1\%. Interestingly, for Socratic LLMs, the best-performing model at world modeling does not translate to the best one in procedural planning. We hypothesize that perception is an important component for models to be able to extend their single-step performance to longer-horizon tasks. Additionally, it can be interpreted as a trade-off between stronger reasoning capabilities without visual grounding using Socratic LLMs, and better perceptual grounding using VLMs but no explicit reasoning.

Video diffusion models exhibit comparatively lower performance, with CogVideoX-I2V reaching 30.1\% and I2VGenXL achieving 26.1\%. These results suggest pixel-space generation struggles to effectively capture detailed action-state causal relationships (diffusion generations are shown in \S\ref{app:action_equivalents_and_samples} and Fig.~\ref{fig:video_diffusion_generation}), and that using better image features (DINOv2 features instead of RGB) for candidate selection does not have much impact on the results. Another limitation of diffusion models is the absence of a reliable method for selecting the correct candidate sequence. Although using the final frames may appear intuitive, it proves ineffective in accurately linking the transition to $\mathcal{O}(s_{t+1})$. For the \textsc{WorldPrediction-PP} task, OEPP-based planners perform at a comparable level with the best zero-shot large models’ performance, while being significantly smaller.

We also analyze the performance of various procedural planners in greater detail in Table~\ref{tab:planning_ablation}. The results reveal a distinct advantage of OEPP models in the in-domain scenarios (COIN and CrossTask), with the PDPP model achieving the highest in-domain accuracy of 49.2\%. However, their out-of-domain performance (EgoExo4D, EPIC-KITCHENS-100, IKEA-ASM) is considerably lower (around 29\%), reflecting a limitation in generalization. Moreover, when provided with oracle captions (human annotations), OEPP performance substantially improves, reaching up to 70.6

\subsection{Human Evaluation and Filtering}

To ensure the quality and robustness of the \textsc{WorldPrediction} benchmark, we conducted a large-scale human evaluation and filtering process. We initially constructed 1,500 samples for both the World Modeling and Procedural Planning tasks. Each sample was then independently solved by two human annotators, following detailed task-specific instructions and example demonstrations. In the World Modeling task, annotators were presented with two context images along with four candidate video actions, and asked to select the action that correctly leads from the initial to the final state. For the Procedural Planning task, annotators were given the context images, a set of video actions, and four possible sequences that order those actions, and were asked to select the correct procedural plan to reach the final state. We adopted a conservative filtering criterion: only samples where both annotators independently provided the correct answer were retained. After filtering, we obtained 825 high-quality samples for \textsc{WorldPrediction-WM} and 570 samples for \textsc{WorldPrediction-PP}, ensuring that human performance was effectively perfect on the released benchmark. Notably, due to the increased complexity of the Procedural Planning task — which requires reasoning over temporally extended sequences rather than single transitions — a smaller proportion of samples was retained.

To maintain a balanced evaluation across plan complexities, we ensured that the number of PP samples was approximately uniform across plan lengths $T$ from 5 to 10, with a higher density of shorter plans (lengths 3 and 4) to reflect their relative frequency and solvability. These human evaluation results underscore the difficulty of our benchmark: in contrast, the best current model performance, Claude-3.5 on \textsc{WorldPrediction-WM}, achieves only 45\% accuracy, with most models ranging between 30–40\% accuracy as shown in Table~\ref{tab:performance_comparison}. For Procedural Planning, even trained planners such as OEPP reach only around 40\% accuracy, and zero-shot frontier models around 37\%, highlighting a significant gap between machine and human performance. Further details regarding the annotation process, including inter-annotator agreement scores, annotation instructions, and annotator workload distribution, are provided in Appendix~\ref{app:human_filtering}.

\section{Conclusion}

In this work, we introduced \textsc{WorldPrediction}, the first benchmark designed to assess high-level world modeling and long-horizon procedural planning from purely visual observations. Unlike prior efforts that focused on low-level physical dynamics or short-horizon tasks, \textsc{WorldPrediction} emphasizes semantic and temporal abstraction, better aligning with the properties of understanding high-level human activities. Evaluations across SOTA VLMs, LLMs, diffusion models, and procedural planning models suggest that world modeling and procedural planning are still two tasks that frontier models largely struggle with, despite humans easily solving both tasks. Current best-performing models largely rely on textual descriptions to tackle both tasks, especially procedural planning, whereas humans are able to solve these tasks from observations alone. Filling this gap is essential for providing models with a better understanding of our world at a higher level and enabling future AI systems to assist humans in a variety of tasks.

\newpage
\bibliographystyle{assets/plainnat}
\bibliography{main}

\clearpage
\newpage
\beginappendix

\section{Additional Information on Human Annotations}
\label{app:human_filtering}

\subsection{Human Annotation Statistics}

In this section, we provide additional information concerning the human evaluation setup. A total of 34 annotators for World Modeling and 46 annotators for Procedural Planning were asked to solve the initial total of 1500 samples for each tasks, while ensuring that each sample will be solved by two different annotator. We ask that annotators should work on a minimum of 20 samples to have time to acclimate themselves to each task, and a maximum of 100 samples to avoid diminishing attention and quality. This resulted in each annotator solving an average of 88 samples for World Modeling, and 65 samples for Procedural Planning, which is effectively more difficult and time-consuming to solve. We provide the inter-annotator agreement on the original split of the benchmark for both tasks in table \ref{tab:iaa_workload}, with 73\% on World Modeling and 65\% on Procedural Planning, showing substantial agreement and reliability of the annotation results.

\begin{table}[h]
  \centering
  \begin{tabular}{lccc}
    \toprule
    Dataset & \# Annotators & Avg. \# Samples per Annotator &  Inter-Annotator Agreement \\
    \midrule
    \textsc{\textsc{WorldPrediction-WM}} & 34 & 88 & 0.73 \\
    \textsc{\textsc{WorldPrediction-PP}} & 46 & 65 & 0.65 \\
    \bottomrule
  \end{tabular}
  \caption{Number of annotators, average number of samples evaluated per annotator and inter-annotator agreement for the human evaluation and filtering.}
  \label{tab:iaa_workload}
\end{table}

\subsection{Human Annotation Setting}

\begin{figure}[H]
  \centering
  \setlength{\abovecaptionskip}{4pt}
  
  \begin{subfigure}{\linewidth}
    \centering
    \includegraphics[width=\linewidth, trim=0 100 0 100]{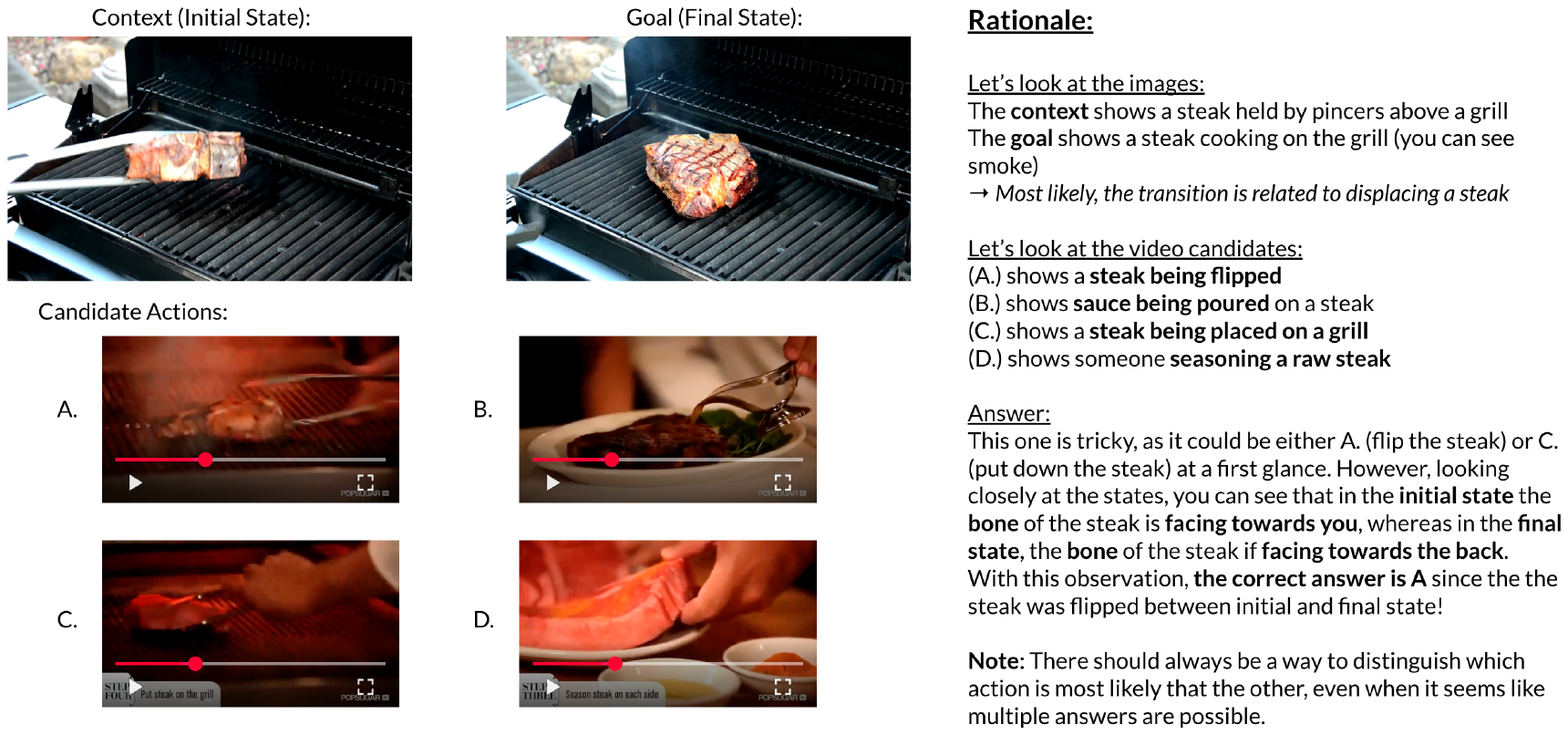}
    \caption{Solved Example with Rationale for the World Modeling task}
    \label{fig:annotation_wm}
  \end{subfigure}
  
  \vspace{20pt}
  
  \begin{subfigure}{\linewidth}
    \centering
    \includegraphics[width=\linewidth, trim=0 110 0 110]{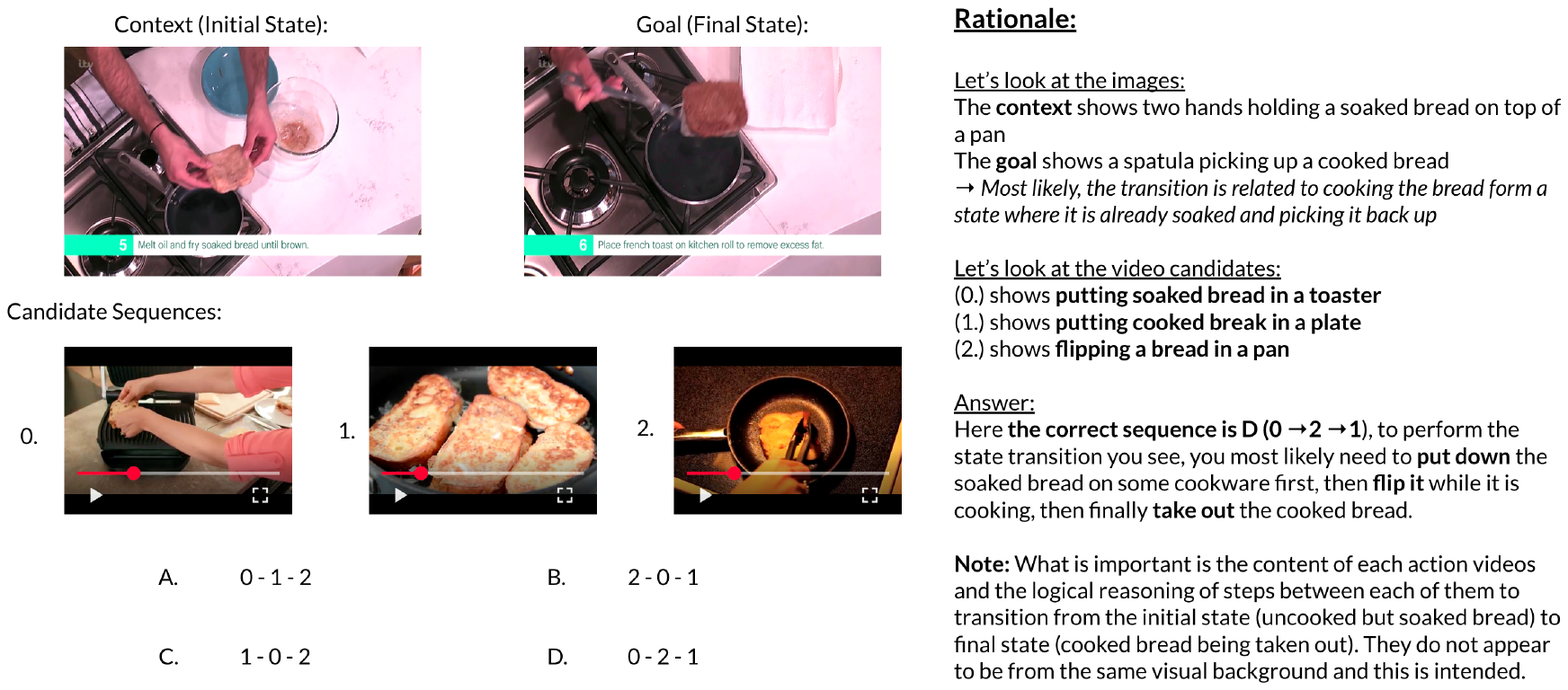}
    \caption{Solved Example with Rationale for the Procedural Planning task}
    \label{fig:annotation_pp}
  \end{subfigure}
  
  \caption{Solved Examples along with correct rationale on how to solve the task for both \textsc{WorldPrediction-WM} and \textsc{WorldPrediction-PP}, provided to the annotators to understand how to evaluate the two tasks.}
  \label{fig:annotation_worldprediction}
\end{figure}

Before starting the annotation task, as the tasks can be conceptually confusing for humans due to the use of action equivalents, each annotator is given four solved examples of World Modeling and two solved examples of procedural planning along with the explanation of how to choose the correct candidate. One solved example for World Modeling is shown in Figure \ref{fig:annotation_wm} and a solved example for Procedural Planning is shown in Figure \ref{fig:annotation_pp}. Along with the solved examples, the annotators are given the following in-depth instructions:

\begin{tcolorbox}[colback=gray!10!white, colframe=gray!80!black, title=World Modeling Instruction for Human Annotation]
\textit{For the World Modeling task, you’ll see two images showing a \textbf{“before”}, as context, and an \textbf{“after”}, as goal, situation (for example, an empty cooking pot as “before”, and a cooking pot containing water as “after”). \textbf{Your job is to select which one of the four provided videos correctly shows the action performed to transition from the first initial state image to the second final state image.}
Please pay attention to the action itself instead of the visual background (scenery or objects). We intentionally sampled the videos to depict the actions performed in a completely different environment (continuing the last example, the correct video answer could be showing a different liquid, like milk, being poured in a different pot: what matters is the performed action itself, here it would have been “Pouring liquid into container”).}
\end{tcolorbox}

\begin{tcolorbox}[colback=gray!10!white, colframe=gray!80!black, title=Procedural Planning Instruction for Human Annotation]
\textit{For the Procedural Planning task, you’ll see two images showing a \textbf{“before”}, as context, and an \textbf{“after”}, as goal, situation (for example, ingredients laid out separately, and then a finished sandwich). \textbf{Your job is to select which one of the provided sequences of videos (each consisting of several short video clips) correctly shows the correct order of action sequence to transition from the first initial state to the second final state image.}
Please pay attention to the actions themselves instead of the visual background (scenery or objects), as we intentionally selected videos depicting the correct actions but performed in completely different environments (continuing the last example, the correct sequence could be something like (1) put the ham on some bread (2) put the cheese (3) close the sandwich, but each action could be depicted in a different environment)}
\end{tcolorbox}

\newpage
\section{Additional Dataset Information}
\label{app:additional_dataset_info}

\begin{figure}[h]
  \centering
  \setlength{\abovecaptionskip}{4pt}

  \begin{subfigure}[b]{0.49\linewidth}
    \centering
    \includegraphics[width=\linewidth, trim=0 50 0 0, clip]{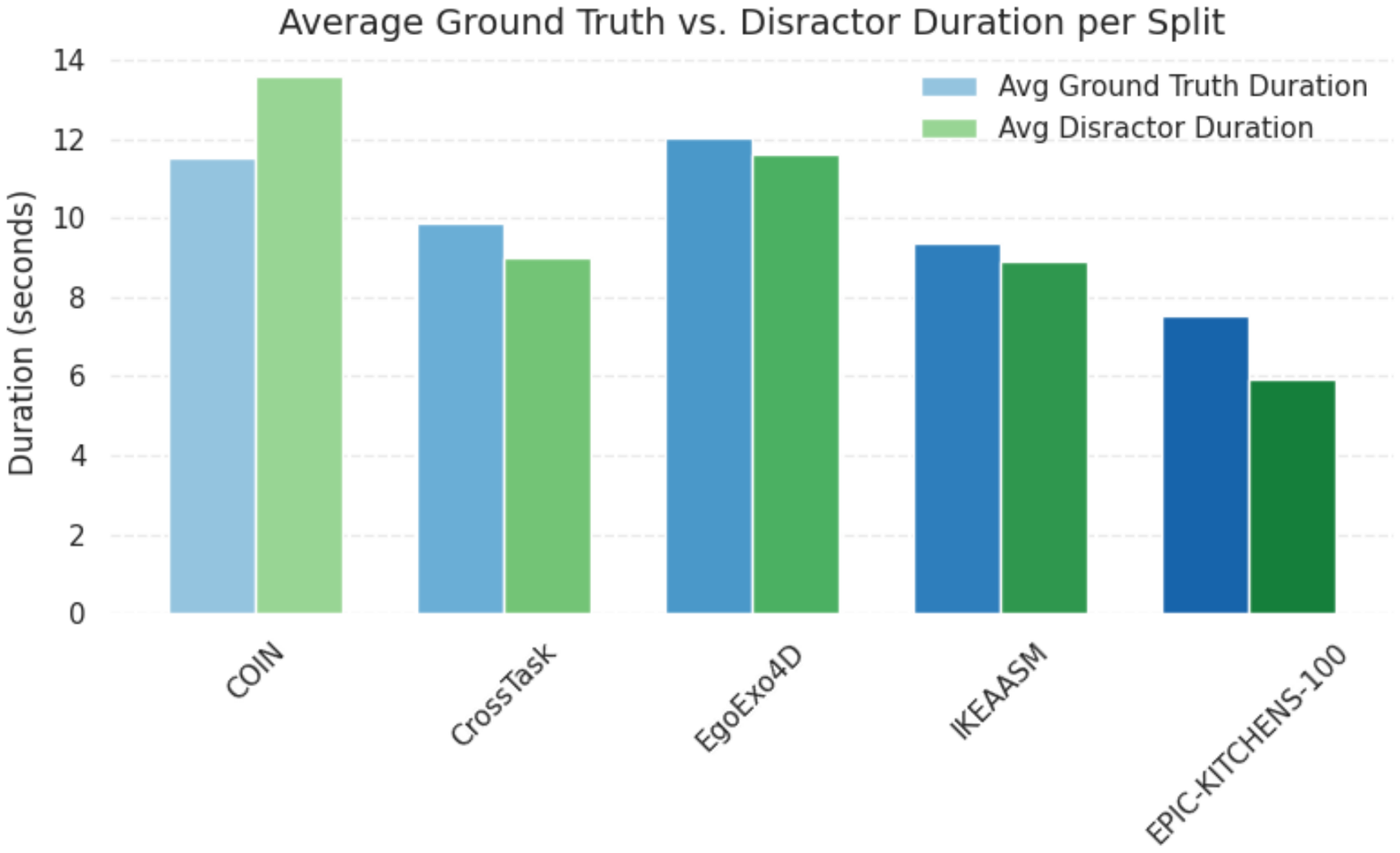}
    \caption{Average duration of action per split (WM)}
    \label{fig:wm_duration_per_split}
  \end{subfigure}\hfill
  \begin{subfigure}[b]{0.49\linewidth}
    \centering
    \includegraphics[width=\linewidth, trim=0 50 0 0, clip]{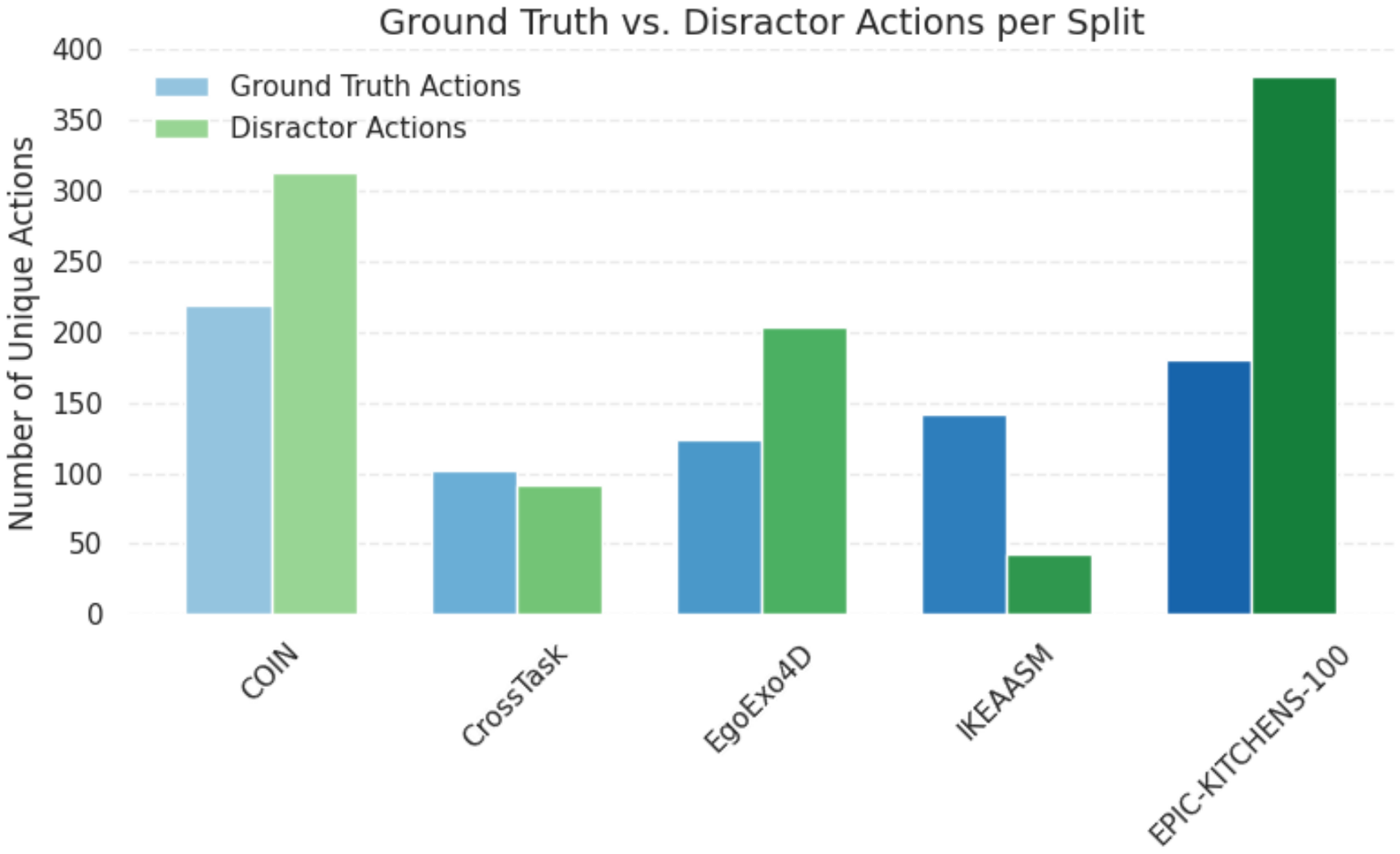}
    \caption{\# unique actions per split (WM)}
    \label{fig:wm_action_per_split}
  \end{subfigure}

  \vspace{12pt} 

  \begin{subfigure}[b]{0.49\linewidth}
    \centering
    \includegraphics[width=\linewidth, trim=0 50 0 0, clip]{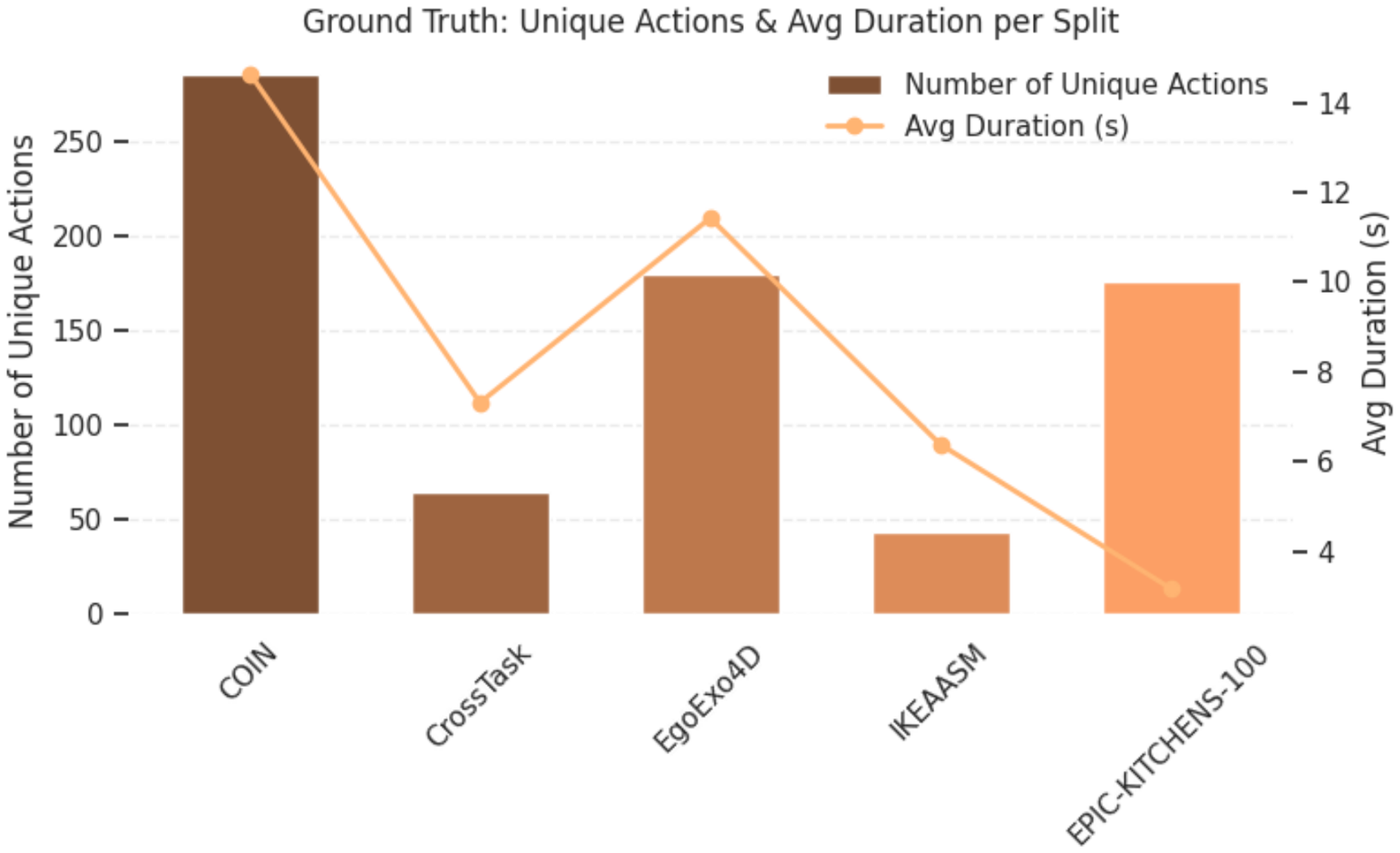}
    \caption{Duration and \# actions per split (PP)}
    \label{fig:pp_duration_and_action_per_split}
  \end{subfigure}\hfill
  \begin{subfigure}[b]{0.49\linewidth}
    \centering
    \includegraphics[width=\linewidth, trim=0 50 0 0, clip]{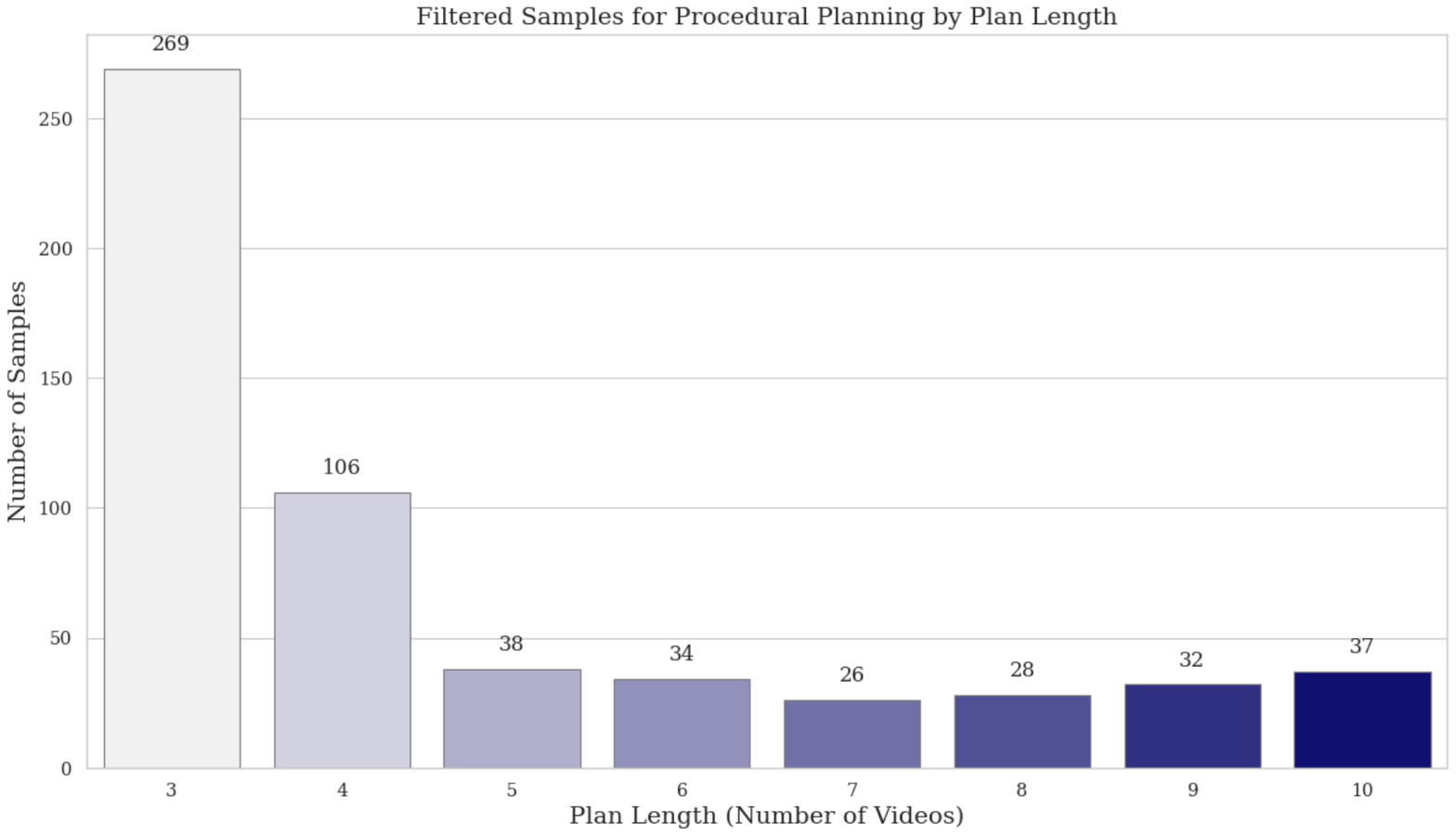}
    \caption{\# Samples per Plan Length (PP)}
    \label{fig:pp_samples_planlength}
  \end{subfigure}

  \caption{Additional dataset information: average duration of actions and number of actions per split for both tasks, and number of samples per plan length in Procedural Planning}
  \label{fig:dataset_info_combined}
\end{figure}

For the \textsc{WorldPrediction-WM} task, we show the average duration of the ground truth action vs the average duration of the distractor actions per dataset split in Figure \ref{fig:wm_duration_per_split} and the number of unique actions that appear as ground truth and as distractors per dataset split in Figure \ref{fig:wm_action_per_split}. Similarly for the \textsc{WorldPrediction-PP} task, as the distractors are shuffled version of the same actions, we directly show the unique actions and average duration per dataset split in Figure \ref{fig:pp_duration_and_action_per_split}. 

EPIC-KITCHENS-100 have relatively shorter action observations for both World Modeling and Procedural Planning, this is expected as the original dataset contain a limited amount of samples but extremely fine-grained annotation of actions (e.g., \textit{pick up, put down, open}) while actions in dataset like COIN and EgoExo4D are more macroscopic (e.g. \textit{add, mix, boil}). This is also interesting for obtaining more robust results on our benchmark, as the duration of the action clips is not standardized and hence does not favor any types of models. 

The number of unique action in IKEAASM and CrossTask is smaller than other datasets for two reasons: first because the number of samples are smaller as shown in \ref{tab:dataset_statistics} due to the human filtering, but also because for IKEAASM for example, the action space is very limited as the dataset only contains four different types of furnitures, so the action overlap is significant. This is not a problem in our benchmark as the assembly domain is proportionally well represented, and some of the CrossTask domains overlap with COIN's domains. Finally, we show the number of samples per plan length in Figure \ref{fig:pp_samples_planlength}, with a majority of plans of length 3 and 4 to reflect current planning datasets, but with a uniform number of samples for plans from 5 to 10 with a bit more than 30 on average.

\begin{figure}[h]
  \centering
  \setlength{\abovecaptionskip}{4pt}
  
  \begin{subfigure}{\linewidth}
    \centering
    \includegraphics[width=\linewidth, trim=0 130 0 130]{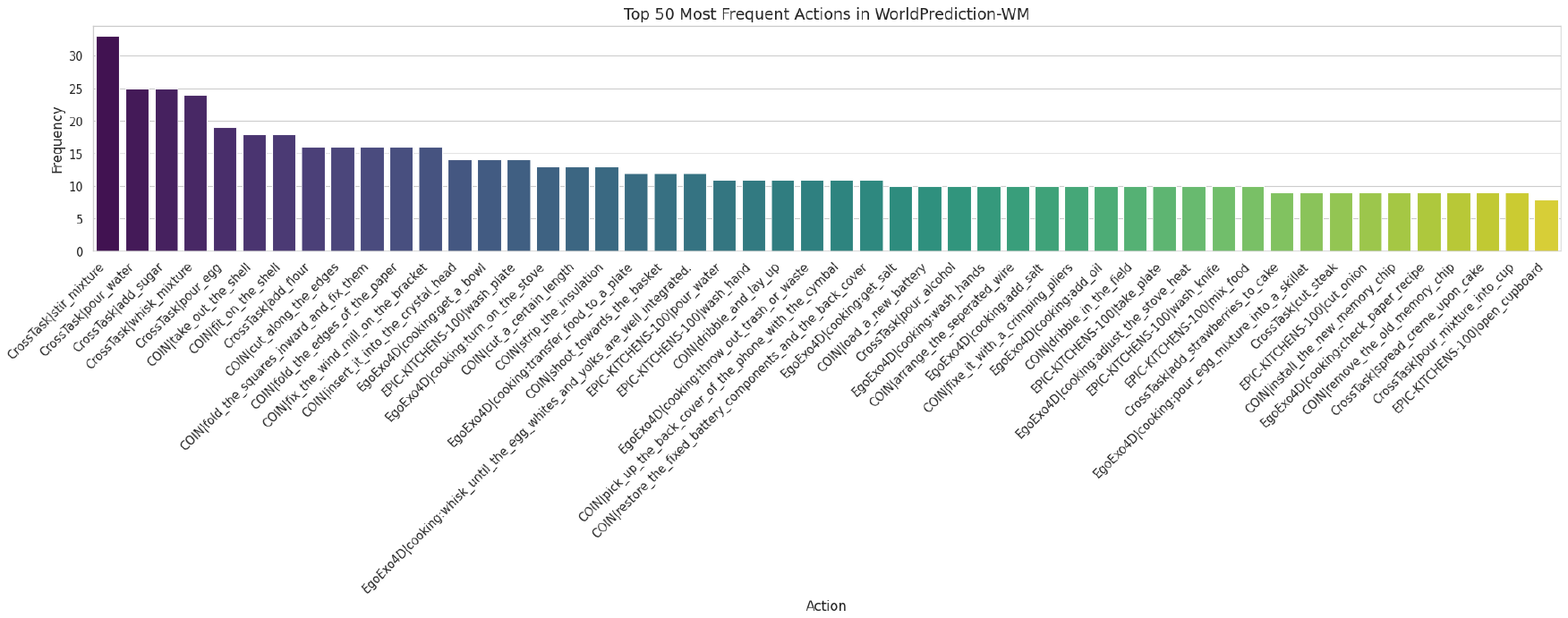}
    \caption{Top-50 Actions appearing in \textsc{WorldPrediction-WM}}
    \label{fig:top50action_wm}
  \end{subfigure}
  
  \vspace{20pt}
  
  \begin{subfigure}{\linewidth}
    \centering
    \includegraphics[width=\linewidth, trim=0 130 0 130]{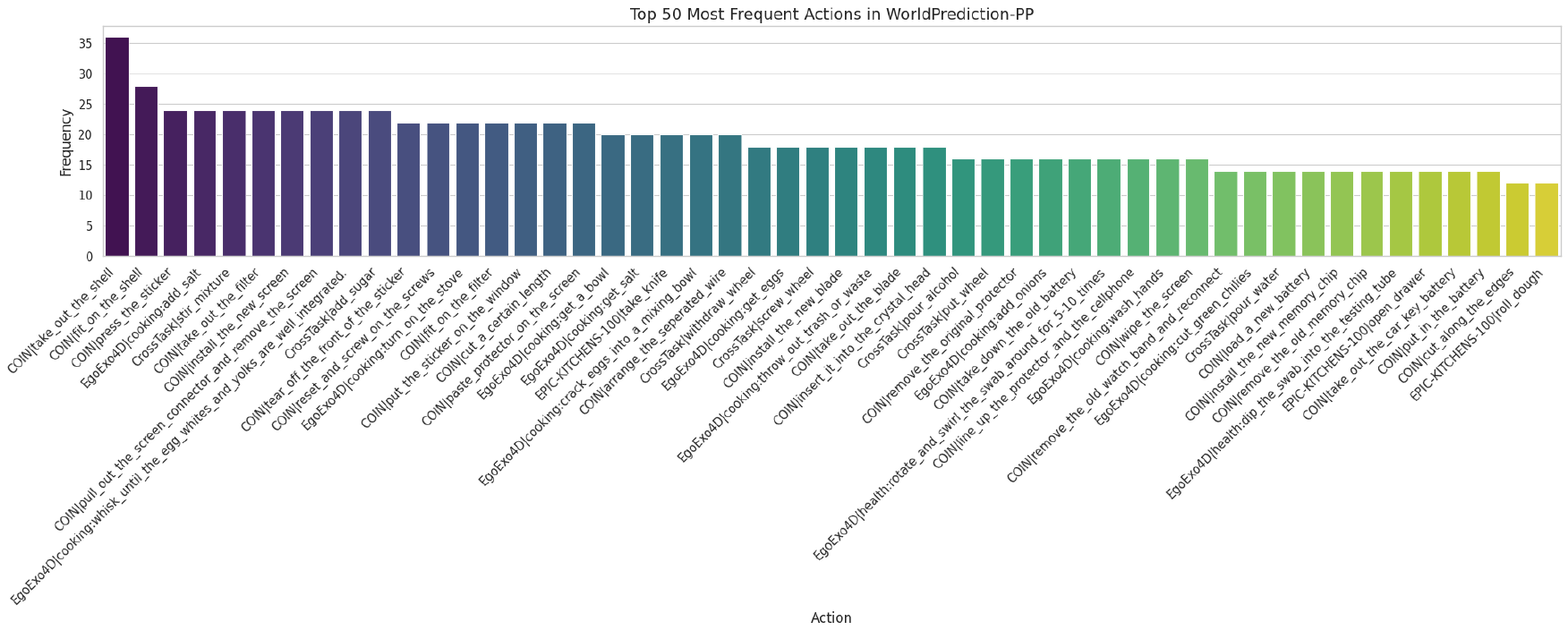}
    \caption{Top-50 Actions appearing in \textsc{WorldPrediction-PP}}
    \label{fig:top50action_pp}
  \end{subfigure}
  
  \caption{Top-50 most frequent actions across \textsc{WorldPrediction-WM} and \textsc{WorldPrediction-PP} datasets (excluding IKEA ASM due to the small action space yielding very high frequency of assembly actions)}
  \label{fig:top50actions_combined}
\end{figure}

We also provide a visualization of the 50 most frequent actions appearing in both the World Modeling and the Procedural Planning tasks in Figure \ref{fig:top50actions_combined}. As the original filtering to deem a World Modeling sample valid vs. a Procedural Planning sample valid differs, the distribution for the action frequency is also different. The action annotations are also provided in the benchmark dataset for researchers interested in only specific domains, tasks or actions. Due to the very small action space of IKEAASM, we choose not to display the actions belonging to the aforementioned split for the figure to be easier to read. The action information concerning IKEAASM can be found on the released dataset benchmark.

\newpage
\section{Action Equivalents}
\label{app:action_equivalents_and_samples}

We show here some of the action equivalents discussed in Section 3.3

\begin{figure}[h]
    \centering
    \includegraphics[width=0.5\linewidth]{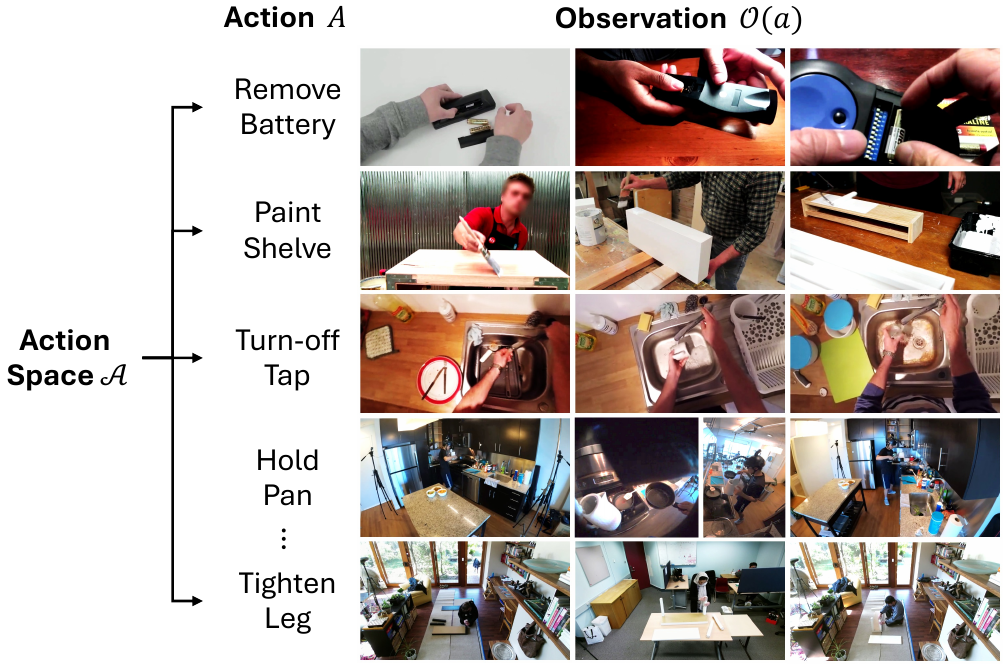}
    \caption{\textbf{High‐level Actions in WorldPrediction.} 
    The action space $\mathcal{A}$ consists of abstract action categories $A$, each instantiated through multiple specific actions $a$ performed across different environments. Each action is represented as a video clip $\mathcal{O}(a)$ (The textual labels are for illustration purposes only and are not included in the benchmark).}
    \vspace{-5pt}
    \label{fig:action_equivalents}
\end{figure}

\end{document}